\documentclass[conference]{IEEEtran}
\usepackage{cite}
\usepackage{amsmath,amssymb,amsfonts}
\usepackage{algorithmic}
\usepackage{graphicx}
\usepackage{textcomp}
\usepackage{xcolor}
\usepackage{booktabs}
\usepackage{multirow}
\def\BibTeX{{\rm B\kern-.05em{\sc i\kern-.025em b}\kern-.08em
    T\kern-.1667em\lower.7ex\hbox{E}\kern-.125emX}}

\ifCLASSINFOpdf
\else
\fi
\begin{document}
%
\title{A Deep Learning Based Ternary Task Classification System Using Gramian Angular Summation Field in \\
fNIRS Neuroimaging Data} 

\author{\IEEEauthorblockN{Sajila D. Wickramaratne}
\IEEEauthorblockA{Electrical and Computer Engineering\\
University of New Hampshire\\
Durham NH, United States\\
Email: sdw1014@wildcats.unh.edu}
\and

\IEEEauthorblockN{Md Shaad Mahmud}
\IEEEauthorblockA{Electrical and Computer Engineering\\
University of New Hampshire\\
Durham NH, United States\\
Email: mdshaad.mahmud@unh.edu}}


%


\maketitle

\begin{abstract}
Functional near-infrared spectroscopy (fNIRS) is a non-invasive, economical method used to study its blood flow pattern. These patterns can be used to classify tasks a subject is performing. Currently, most of the classification systems use simple machine learning solutions for the classification of tasks. These conventional machine learning methods, which are easier to implement and interpret, usually suffer from low accuracy and undergo a complex preprocessing phase before network training. The proposed method converts the raw fNIRS time series data into an image using Gramian Angular Summation Field. A Deep Convolutional Neural Network (CNN) based architecture is then used for task classification, including mental arithmetic, motor imagery, and idle state.  Further, this method can eliminate the feature selection stage, which affects the traditional classifiers' performance. This system obtained 87.14\% average classification accuracy higher than any other method for the dataset.
\end{abstract}


%
\IEEEpeerreviewmaketitle
\section{Introduction}

Functional near-infrared spectroscopy is a practical, optical neuroimaging method that monitors the hemodynamic response to brain activation. Specific parameters of the response monitored with fNIRS comprise changes in oxygenated and deoxygenated hemoglobin at particular loci across the cortical surfaces evoked following stimulation\cite{leff2011assessment}. Brain-computer interfaces have emerged as a novel mode of communication for individuals who have lost voluntary movements \cite{blankertz2016berlin}. Various fNIRS experiments for (BCI) and Human-Machine Interaction (HMI) applications have been investigated task classification, including cognitive tasks and motor tasks \cite{naseer2015fnirs}.Support Vector Machines (SVM) and Linear Discriminant Analysis (LDA) based methods are frequently reported as the highest accuracy classifiers. However, performing classification using conventional methods such as SVM requires a prior feature selection and preprocessing.

Deep Learning (DL) can overcome the challenge of feature selection, as it extracts features directly from the fNIRS signal and requires minimum feature preprocessing. However, DL for fNIRS-based classification has been applied in only a few studies. Bashivan et al. introduced a method to map the electrodes' location to a 2D plane, building a topological image structure for representing EEG\cite{bashivan2016mental}. An improvement in classification was observed relative to SVM by Saadati et al., who used Deep Neural Network (DNN) and CNN with a temporal framework for image construction for various cognitive and motor tasks\cite{saadati2019convolutional}.

 This paper proposes using a multi-class fNIRS-based system that classifies three brain activation patterns recorded during motor imagery, mental arithmetic, and idle state with a deep learning algorithm. Although traditional machine learning methods are used frequently for fNIRS based task classification, in recent studies, deep learning classifiers have obtained satisfactory results\cite{tanveer2019enhanced} \cite{chiarelli2018deep}. The proposed system uses a Convolutional Neural Network system to classify Gramian Angular Summation Field images generated using only fNIRS data. It has a classification accuracy of 87.14\%, a significant improvement from the original classifier, which used both fNIRS and EEG data.

\section{Method}
In this study, a CNN based deep learning model was developed for ternary classification.  Further analysis was also done using this data to compare the performance of models that use different feature sets and even raw data. Deep Learning classifiers are potent tools that can handle data with complex nonlinear relationships; hence, the classification accuracy into varying data preparation levels can bring insight into how much data preprocessing is required for the deep learning classifiers.
\subsection{System Overview}
\begin{figure*}[h]
        \centering
        \includegraphics[scale=0.5]{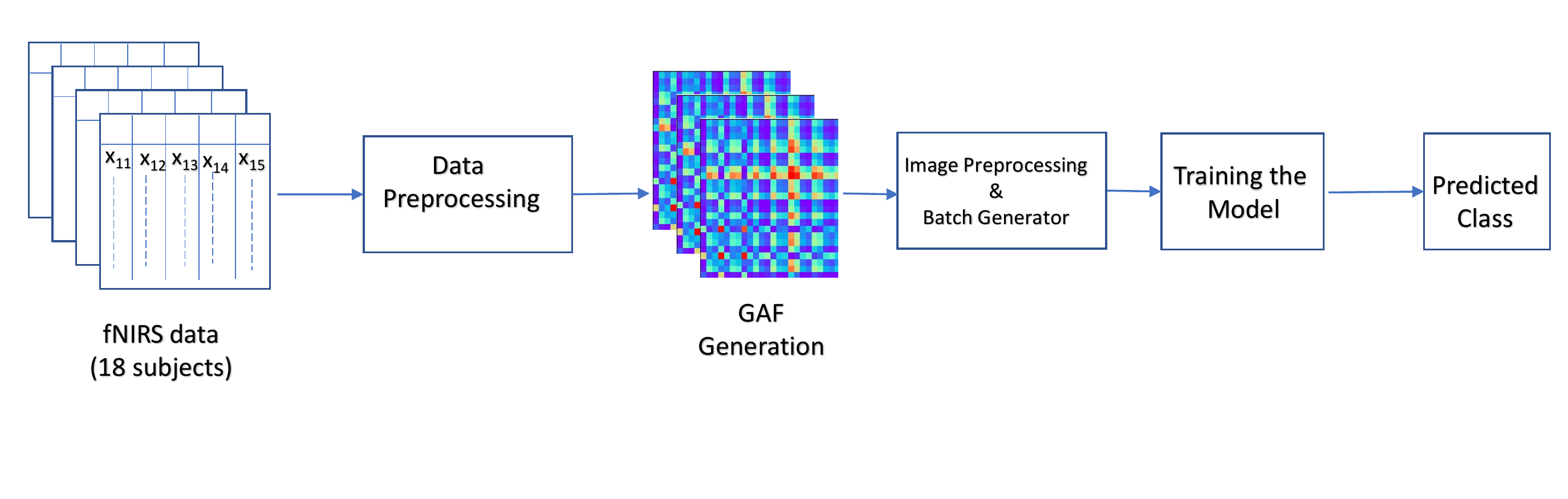}
        \caption{System overview for the CNN based Prediction System }
        \label{fig:sys_overview}
    \end{figure*}

The fNIRS data were analyzed in several different ways. The Fig.\ref{fig:sys_overview} illustrates the Overview of the complete system. The individual components of the system will be described in detail in the later sections. The acquired data is preprocessed in the first step. This preprocessed data is then used to generate the images representing a pre-chosen channel's time series data. These images are then preprocessed and arranged in batches. These batches are used to feed to a deep learning network trained on only raw channel data. Other approaches ware used to prepare data for the classifiers used for comparison with the proposed model.  After creating batches, the data is fed to the model to be trained. Finally, the result is obtained with the task being classified into MA, MI, or IS.

    
\subsection{Data}
The data was obtained from the study conducted by Shin et al. \cite{shin2016open}. fNIRS data were collected using a portable NIRS system at a sampling rate of 13.3 Hz with 16 NIRS channels for 18 subjects. For the right-hand MI task, the subjects performed an intricate finger-tapping pattern at approximately 2 Hz. For the MA task, the subjects had to continuously subtract a one-digit number from the result of a former calculation as fast as possible. Each of the tasks was performed by the subjects 30 times in random order.

\subsection{Pre-Processing}
The detected optical densities (ODs) of fNIRS data were converted to hemodynamic variations by implementing the formula proposed by Matcher et al. .\cite{matcher1995performance}. The converted HbR and HbO values were band-pass filtered using a 3rd-order Butterworth zero-phase filter with a passband of 0.01–0.09 Hz. NIRS data were segmented into epochs from -5 to 25~s.  By subtracting the temporal mean value between 1 and 0 s from each NIRS epoch baseline correction was performed.

\begin{figure}[h]
        \centering
        \includegraphics[scale=0.35]{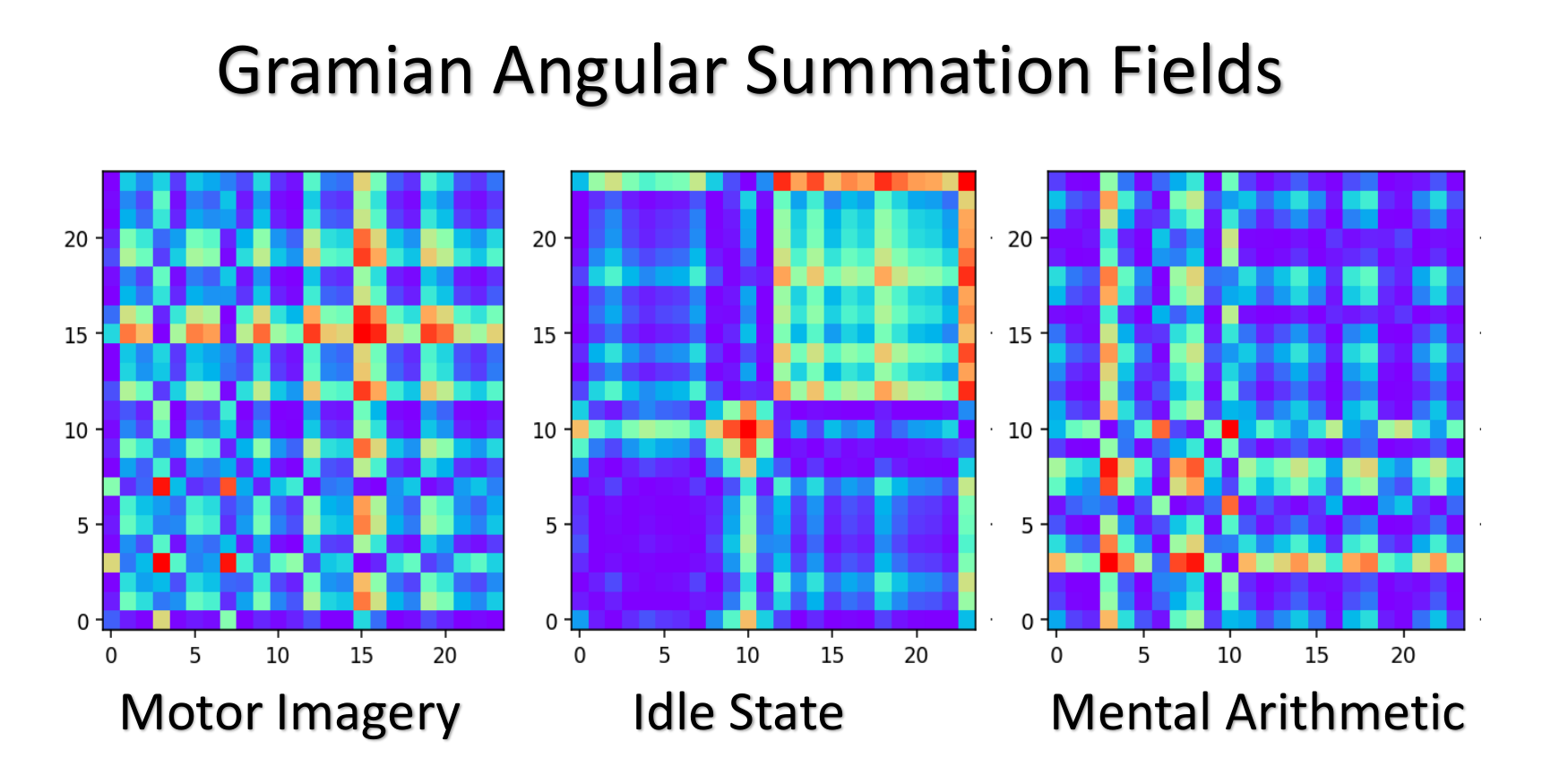}
        \caption{The Gramian Angular Fields for Motor Imagery, Idle State and Mental Arithmetic tasks}
        \label{fig:gram_maps}
    \end{figure}

 A Gramian Angular Field(GAF) is an image obtained from a 1-dimensional time series, representing some temporal correlation between each time point. Two methods are available: Gramian Angular Summation Field(GASF) and Gramian Angular Difference Field\cite{wang2015imaging}. A GAF, in which we represent time series in a polar coordinate system instead of the typical Cartesian coordinates. For this study, GASF images were used as the images to train the Convolution Neural Networks. After analyzing the best performing baseline classifier's feature importance, the fNIRS channel used to generate the images was chosen. The GASF images obtained for MI, MA, and IS tasks are shown in Fig.~\ref{fig:gram_maps}.


\subsection{Feature Extraction and Selection}
The proposed CNN prediction model does not need specific Feature Extraction and Selection methodology. For the traditional classifiers used, ANN and Bi-LSTM models used for comparison required feature selection due to varying degrees. The feature extraction and feature selection were made in several steps. For comparison, several different feature sets were extracted in the initial phase.NIRS feature vectors were constructed using the temporal mean values of HbR and HbO in the 5-10~s and 10-15~s temporal windows in NIRS epochs from all channels, considering the hemodynamic delay. These features were used to train the traditional classifiers and ANN classifier. For the Bi-Directional LSTM classifier, the raw data passed through a dimension reduction methodology. For this instance, a kernal Principal Component Analysis was used, from which 20 components were selected based on the explained variance.

\section{Model}
\begin{figure*}[h]
        \centering
        \includegraphics[scale=0.5]{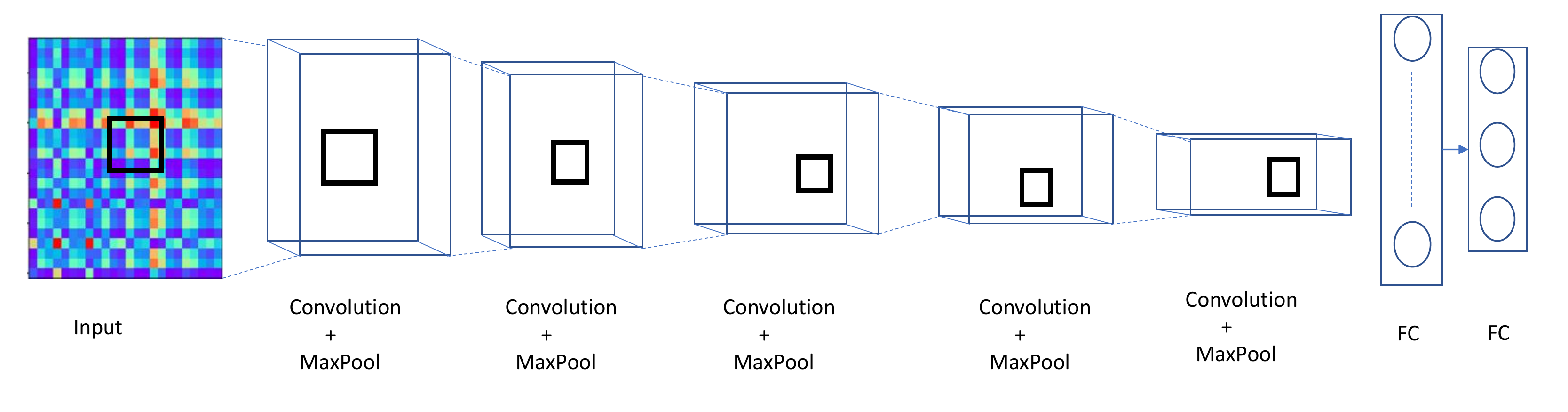}
        \caption{The Proposed Convolution Neural Network}
        \label{fig:cnn_mod}
    \end{figure*}
CNN are predominantly used in computer vision tasks like image classification. The operation of CNN can be described in 3 phases. The first phase is the convolution. A feature map is created by scanning a few pixels each time of an image. Along with the feature map the probabilities that each feature belongs to a specific class is also generated. The second phase is pooling and reducing dimensionality while retaining the most relevant information is the main purpose of pooling. An overview of the critical elements in the image is created by pooling. Max Pooling is used frequently for CNNs, where the highest value is taken from each area scanned by the CNN. A CNN typically has to perform multiple turns of convolution and pooling.

 This fully-connected neural network analyzes the final probabilities and decides which class the image belongs to. The fully connected layer(s) perform classification on the extracted features based on information in labeled training data. Every node in a fully connected layer is connected to every node in the previous layer. Finally, the output layer contains a single node for each target class in the model with a softmax activation function to compute each class's probability. The softmax activation function ensures that the final outputs fulfill the constraints of a probability density.

As shown in Figure\ref{fig:cnn_mod}, the CNN model consisted of 18 layers, including an input layer, four pairs of convolutional, max-pooling layers, Batch Normalization, two fully connected layers, a Dropout Layer, a softmax layer, and a classification output layer. All the Convolution Layers were activated by the Rectified Linear Unit(Relu) function.

Dropout is applied to all hidden layers' output, and all layers have l2-kernel regularizer of strength 0.3. The model was trained using batches of 8, which were generated from an image generator. The learning rate was reduced on the plateau, and early stopping was used to reduce overfitting. For all the models, the loss function used was categorical cross-entropy, and the optimizer was  RMSprop.

\section{Results}
The results of the study are divided into two sections. The first section will focus on the baseline classifiers that were used to evaluate the data set. The results from these classifiers were used to determine some parameters for the deep learning classifiers. The second section will present the proposed deep neural network's performance and its comparison with the other classifiers.

Area under Receiver-Operator Characteristic (AUROC) is an important metric to determine the classifier's ability to distinguish between the classes accurately. The precision-recall curve illustrates the trade-off between precision and recall considering different thresholds. A high area under the curve translates to both high recall and high precision. High scores for both show that the classifier is returning accurate results, as well as returning a majority of all positive results.
\begin{figure}[h]
        \centering
        \includegraphics[scale=0.45]{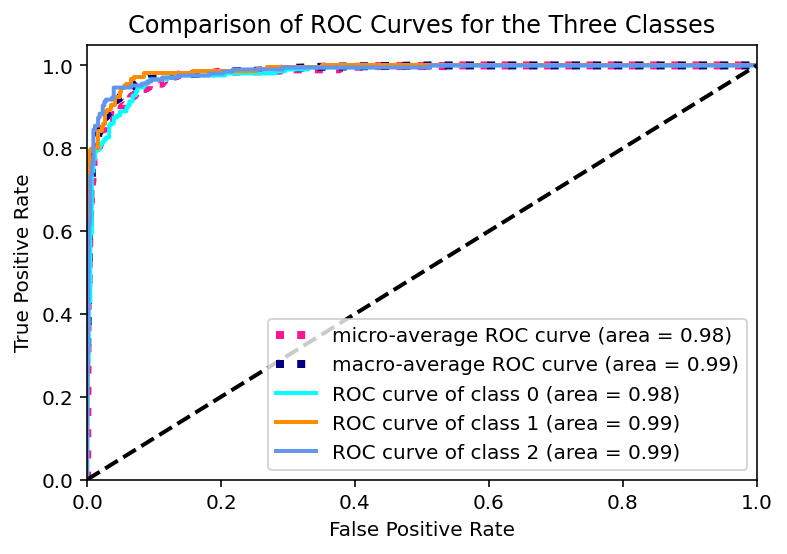}
        \caption{The Comparison for Receiver Operater Characteristic for the three classes using test data from subject 16}
        \label{fig:roc_cnn}
\end{figure}

\subsection{Baseline Classifier Performance}
In the first part of the analysis, the traditional classifiers were used to evaluate the data set. A 10 fold cross-validation methodology was used for the evaluation. The results obtained from this analysis are given in Table\ref{tab:baseline_class}. From the considered classifiers, XGBoost performed best considering both classification accuracy(58.0\%) and AUROC(0.7022).

Due to their ease of interpretation, the baseline classifiers were used to find the features that affect the result. It was achieved by comparing the mean SHAP value obtained from the Variable Importance Plot for each feature corresponding to the XGBoost classifier\cite{NIPS2017_7062}. A variable importance plot lists the most significant variables in descending order. From the variable importance plot, jumping means for the [10-15]s time interval of the epoch for channel 17 from the original data set seems to be the most significant variable. This channel was chosen as the univariate time series to generate GASF images to train the proposed CNN network.
\subsection{Deep Learning Classifier}
The original data classifier used by Shin et al. used  10-fold cross-validation with shrinkage linear discriminant analysis (sLDA) for each of the three binary classification problems. This sLDA based system obtained an average classification accuracy of 66.08\%. This was compared with multi-layer Artificial Neural Networks, Bi-LSTM Networks, and the proposed CNN network. The subject-wise performance comparison of these classifiers is given in Table~\ref{tab:class_acc}. The proposed model has different AUCs for different tasks. The AUROC values vary across the subjects as well. Fig.~ \ref{fig:roc_cnn} shows the ROC curve for the CNN model trained with subject 16 data. The tasks 0,1, and 2 correspond to the MI, MA, and IS tasks, respectively. The micro-average for the ROC area under all three classes is 0.98.  The average Precision score micro averaged over all the classes was 0.97 was obtained.
\begin{table}[]
\caption{Comparison Between the Classifiers}
\label{tab:baseline_class}
\begin{tabular}{|c|c|c|}
\hline
\textbf{Model}      & \textbf{Classification Accuracy\%} & \textbf{AUROC} \\ \hline
Logistic Regression & 44.62                           & 0.6230         \\ \hline
KNN                 & 40.49                           & 0.5782         \\ \hline
Random Forest       & 42.83                           & 0.6160         \\ \hline
SVM                 & 40.74                           & 0.5957         \\ \hline
XGBoost             & 58.02                           & 0.7022         \\ \hline
\end{tabular}
\end{table}

\begin{table}[]
\centering
\caption{Comparison Between the Classification Accuracy of the Classifiers for each Subject}
\label{tab:class_acc}
\begin{tabular}{|c|c|c|c|c|}
\hline
\multirow{2}{*}{\textbf{\begin{tabular}[c]{@{}c@{}}Subject\\ No\end{tabular}}} & \multicolumn{4}{c|}{\textbf{Classification Accuracy\%}}                   \\ \cline{2-5} 
                                                                               & \textbf{sLDA}  & \textbf{ANN}   & \multicolumn{1}{l|}{\textbf{Bi-LSTM}} & \multicolumn{1}{l|}{\textbf{CNN}} \\ \hline
\textbf{1}                                                                     & 73.11 & 75.55 & 85.55                        & 82.72                    \\ \hline
\textbf{2}                                                                     & 68.33 & 62.26 & 83.26                        & 84.31                    \\ \hline
\textbf{3}                                                                     & 54.88 & 58.88 & 72.88                        & 82.75                    \\ \hline
\textbf{4}                                                                     & 79.22 & 67.77 & 90.11                        & 84.44                    \\ \hline
\textbf{5}                                                                     & 52.11 & 54.44 & 66.44                        & 87.62                    \\ \hline
\textbf{6}                                                                     & 75.44 & 68.88 & 88.88                        & 83.94                    \\ \hline
\textbf{7}                                                                     & 49.88 & 46.23 & 66.23                        & 82.35                    \\ \hline
\textbf{8}                                                                     & 57.77 & 56.66 & 75.44                        & 90.02                    \\ \hline
\textbf{9}                                                                     & 62.22 & 63.51 & 81.43                        & 89.37                    \\ \hline
\textbf{10}                                                                    & 68.22 & 62.61 & 80.45                        & 90.45                    \\ \hline
\textbf{11}                                                                    & 67.11 & 63.91 & 80.91                        & 89.72                    \\ \hline
\textbf{12}                                                                    & 65.44 & 77.23 & 77.23                        & 88.02                    \\ \hline
\textbf{13}                                                                    & 55.00 & 50.21 & 68.21                        & 88.10                    \\ \hline
\textbf{14}                                                                    & 78.77 & 72.11 & 88.11                        & 90.82                    \\ \hline
\textbf{15}                                                                    & 67.66 & 70.23 & 82.23                        & 87.58                    \\ \hline
\textbf{16}                                                                    & 88.77 & 80.00 & 96.01                        & 93.13                    \\ \hline
\textbf{17}                                                                    & 59.33 & 62.22 & 72.21                        & 85.28                    \\ \hline
\textbf{18}                                                                    & 67.44 & 59.66 & 82.12                        & 87.97                    \\ \hline
\textbf{Average}                                                               & \textbf{66.08} & \textbf{64.02} & \textbf{79.88}                        & \textbf{87.14 }                   \\ \hline
\end{tabular}
\end{table}

\section{Discussion}
The results of our study show that fNIRS based classification of tasks can be increased by using deep learning methods. This approach can be an essential step to enhance fNIRS based classification systems. The proposed ternary classification system can classify MA, MI, and IS-related brain activation patterns. Some form of preprocessing was required for all the methods used in this paper. However, the feature selection step can be eliminated by using CNN networks.

The MI task is associated with activating the mid-region of the frontal lobe. The fNIRS data were not collected from that area due to practical difficulties with the setup. Some valuable information is lost by not being able to acquire data from that region. In the original publication, which used this data set by Shin et al. , EEG data were used to overcome this problem.\cite{shin2018ternary}. In the proposed system, the classification accuracy of 87.14\% is higher than the accuracy of the hybrid fNIRS-EEG by Shin et al. Hence this system has the potential to improve even further if the EEG data can be integrated into the system. Since only fNIRS data were used for the proposed system's training, one way that accuracy can be improved is by using several fNIRS data channels to generate the images. Hence further research is required into the mentioned improvement.

\section{Conclusion}
This paper presented a comparison between different classification systems that were trained using fNIRS data. Brain imaging data are gaining popularity with extensive research on BCI interfaces. Traditionally, simpler classification systems such as SVM or LDR were used for classification tasks. The method proposed CNN network obtained a classification accuracy of 87.14\%. The proposed method improved over the previous ternary classification system, which used both EEG and fNIRS data. The proposed model can be enhanced by integrating other modalities like EEG data. In this study, we used the traditional interpretable classifiers to analyze what parameters affect the outcome most and used this information to enhance the deep-learning classifier's performance. Further studies should be conducted on how several time series data streams can be represented in an image format. This will enable integrating other modalities to the proposed CNN classification system, which can enhance the performance.





%
\bibliographystyle{IEEEtran}
\bibliography{fnirs}

\end{document}